\documentclass{ieeeaccess}
\usepackage{cite}
\usepackage{amsmath,amssymb,amsfonts}
\usepackage{algorithmic}
\usepackage{graphicx}
\usepackage{textcomp}

\makeatletter
\AtBeginDocument{\DeclareMathVersion{bold}
\SetSymbolFont{operators}{bold}{T1}{times}{b}{n}
\SetSymbolFont{NewLetters}{bold}{T1}{times}{b}{it}
\SetMathAlphabet{\mathrm}{bold}{T1}{times}{b}{n}
\SetMathAlphabet{\mathit}{bold}{T1}{times}{b}{it}
\SetMathAlphabet{\mathbf}{bold}{T1}{times}{b}{n}
\SetMathAlphabet{\mathtt}{bold}{OT1}{pcr}{b}{n}
\SetSymbolFont{symbols}{bold}{OMS}{cmsy}{b}{n}
\renewcommand\boldmath{\@nomath\boldmath\mathversion{bold}}}
\makeatother

\usepackage{url}
\usepackage{hyperref}
\usepackage{cleveref}
\usepackage{multirow}
\usepackage{multicol}
\usepackage{bbm}
\usepackage{acronym}
\usepackage{mathtools}

\DeclareMathOperator*{\argmax}{argmax}

\DeclareMathOperator*{\argtopk}{argTopK}

\def\pmm{$\pm$}
\newacro{knn}[$k$NN]{$k$-nearest neighbor}
\newacro{llm}[LLM]{Large Language Model}
\newacro{plm}[PLM]{Pretrained Language Model}
\newacro{nlp}[NLP]{Natural Language Processing}
\newacro{icl}[ICL]{In-Context Learning}
\newacro{rag}[RAG]{Retrieval-Augmented Generation}

\newcommand{\emphname}[1]{\textsc{#1}}

\crefname{figure}{Figure}{Figures}
\crefname{table}{Table}{Tables}
\crefname{algorithm}{Algorithm}{Algorithms}
\crefname{section}{Section}{Sections}
\crefname{subsection}{Section}{Sections}
\crefname{subsubsection}{Section}{Sections}
\crefname{equation}{Equation}{Equations}

\begin{document}
\history{Date of publication xxxx 00, 2024, date of current version Oct. 15, 2024.}
\doi{10.1109/ACCESS.2024.0429000}

\title{End-to-End Trainable Retrieval-Augmented Generation for Relation Extraction}

\author{\uppercase{Kohei Makino}, 
\uppercase{Makoto Miwa}, and \uppercase{Yutaka Sasaki}}
\address{Toyota Technological Institute, 2-12-1 Hisakata, Tempaku-ku, Nagoya, 468-8511 Japan}

\markboth
{Makino \headeretal: End-to-End Trainable Retrieval-Augmented Generation for Relation Extraction}
{Makino \headeretal: End-to-End Trainable Retrieval-Augmented Generation for Relation Extraction}

\corresp{Corresponding author: Yutaka Sasaki (e-mail: yutaka.sasaki@toyota-ti.ac.jp).}

\begin{abstract}
This paper addresses a crucial challenge in retrieval-augmented generation-based relation extractors; the end-to-end training is not applicable to conventional retrieval-augmented generation due to the non-differentiable nature of instance retrieval. This problem prevents the instance retrievers from being optimized for the relation extraction task, and conventionally it must be trained with an objective different from that for relation extraction. To address this issue, we propose a novel End-to-end Trainable Retrieval-Augmented Generation (ETRAG), which allows end-to-end optimization of the entire model, including the retriever, for the relation extraction objective by utilizing a differentiable selection of the $k$ nearest instances. We evaluate the relation extraction performance of ETRAG on the TACRED dataset, which is a standard benchmark for relation extraction. ETRAG demonstrates consistent improvements against the baseline model as retrieved instances are added. Furthermore, the analysis of instances retrieved by the end-to-end trained retriever confirms that the retrieved instances contain common relation labels or entities with the query and are specialized for the target task. Our findings provide a promising foundation for future research on retrieval-augmented generation and the broader applications of text generation in Natural Language Processing.
\end{abstract}


\titlepgskip=-21pt

\maketitle
\section{Introduction}

Relation extraction is a fundamental task in \ac{nlp}, which involves identifying and classifying semantic relationships between entity mentions, such as people, organization, and location names, in text~\cite{chinchor-1998-relation-extraction-first}. Relation extraction plays a crucial role in understanding and interpreting the underlying meaning of sentences by analyzing how entities are interrelated~\cite{tjong-kim-sang-de-meulder-2003-introduction-conll-2003,wang-etal-2016-relation-sentence}.
\footnote{This work has been submitted to the IEEE for possible publication. Copyright may be transferred without notice, after which this version may no longer be accessible.}

Relation extraction is used in practical applications in several domains. For instance, in knowledge graph construction, relation extraction aids in transforming unstructured text into structured data, which can then be used to populate and enrich knowledge graphs~\cite{du-2022-re-for-kg}. In domain-specific scenarios, such as biomedical text mining~\cite{davis-2008-ctd-dataset}, it can extract relationships between genes, diseases, and drugs, which are helpful for advanced research and discovery such as search systems~\cite{zhu-etal-2023-descriptive-bio-search} and prediction of novel things~\cite{long-2022-bio-link-prediction}. Thus, relation extraction is used for a wide variety of purposes.

Developing relation extractors is an ongoing endeavor in \ac{nlp}. Relation extractors are designed to accurately identify and classify relations in text, which requires understanding the nuanced and sometimes complex language structures. Advances in machine learning and \ac{nlp} methods have shaped the evolution of the extractors. Recent extractors are based on deep learning models to obtain high-performance~\cite{miwa-bansal-2016-end,nan-etal-2020-reasoning-lsr,ma-etal-2023-dreeam}, while traditional extractors are rule-based models~\cite{riloff-1993-rule-based-relation-extraction} and feature-based models~\cite{zelenko2003kernel}.

The \acp{plm}~\cite{peters-etal-2018-elmo,devlin-etal-2019-bert,liu-2019-roberta} have become a de facto standard in recent \ac{nlp}, fundamentally changing the field landscape. \Acp{plm} are neural network models pretrained on a common \ac{nlp} task, such as language modeling~\cite{radford-2019-gpt-2} and masked language modeling~\cite{devlin-etal-2019-bert}, and are used by fine-tuning it to fit a target task.
Many studies on relation extraction with \acp{plm} have been conducted because of the high performance~\cite{wang-etal-2022-deepstruct-plm-relation-extraction,huang-etal-2022-unified-re-unist}.

With the advent of \acp{plm}, well-trained text generation models, including \acp{llm}~\cite{brown-etal-2020-gpt3,chung-2022-flan-t5,touvron-2023-llama-arxiv} trained on a larger corpus with larger-scale parameters, are attracting attention~\cite{radford-2019-gpt-2,raffael-et-al-2020-t5,brown-2020-gpt-3}.
Text generation model-based relation extractors~\cite{cohen-2021-relation-two-way-sp,wadhwa-etal-2023-revisiting-llm-relation-extraction,wan-etal-2023-gpt-re} are used because the pretraining and the fine-tuning are similar and it helps the extractor training. These models have been adapted to treat relation extraction as a question-answering task~\cite{cohen-2021-relation-two-way-sp}, a text summarization task~\cite{lu-etal-2022-summarization-sure}, or a language modeling~\cite{wadhwa-etal-2023-revisiting-llm-relation-extraction}, leveraging their knowledge and understanding of natural language.

Some existing works utilize relation instances, which consist of text and relation between entities, to improve the performance. Instance-based methods perform better in low-resource settings with insufficient training data because they can use known examples as anchors. A typical instance-based extractor using \ac{plm} predicts the label of the nearest known instance on the encoded representation~\cite{wan-etal-2022-rescue-nearest-neighbor-relation-extraction}.
However, the inference with simple voting with nearest neighbor methods is weak in terms of the overall performance compared to other models.

Combining text generation models with instance utilization has been particularly effective with \ac{icl}~\cite{brown-etal-2020-gpt3}. In relation extraction via \ac{icl}, instructions and few-shot instances verbalized as text are used as prompts to supervise text generation models from context, and the model generates the text convertible to relation labels conditioned by the prompt~\cite{wadhwa-etal-2023-revisiting-llm-relation-extraction}. \ac{icl}, which predicts based on context, can produce output using instances as hints rather than methods such as the nearest neighbor method.
Since standard \ac{icl} uses pre-defined instances in the prompt, \ac{icl} methods cannot access instances adaptable to input.

To solve these problems, \ac{rag}~\cite{patric-2020-rag-retrieval-augumented-generation} introduces the instances relevant to the input into the text prompt instead of pre-defined instances. In \ac{rag}-based relation extractors~\cite{wan-etal-2023-gpt-re}, retrievers prepared separately from the base model select instances.
The retrievers use the target text as a query and select instances from the relation instances stored as a database. The selected instances are then introduced into the text prompt.
Generally, the retrieval is done by embedding the inputs and instances into a common dense feature space by a separately trained encoder model and selecting the nearest neighbors.
Thus, \ac{rag} introduces adaptive instances to the prompt of the text generation model.

The retriever used in \ac{rag} requires training to retrieve the appropriate instance when used in the target task. 
However, the instance retriever needs to be trained separately from the target relation extraction task because the text generation model with the retriever cannot be trained end-to-end. In other words, designing a retriever for each target problem is necessary, which increases the development cost~\cite{wan-etal-2023-gpt-re}.
Therefore, end-to-end training allows direct optimization of the retriever to the target task, reducing development costs by eliminating the need to devise case-specific training methods.

Our goal is to create an environment where the entire model with \ac{rag} can be optimized for the relation extraction task.
Therefore, this study aims to make the \ac{rag} model end-to-end trainable.
Specifically, we eliminate indifferentiable operations in the retriever that prevent the training of deep learning models by replacing them with differentiable operations.
By integrating the operations, we propose ETRAG (End-to-end Trainable Retrieval-Augmented Retrieval) as a \ac{rag} that can train the entire model end-to-end.
Our proposed method is evaluated by a relation extraction task to confirm effectiveness and characteristics.

Contributions in this paper are threefold.
\begin{itemize}
    \item We propose end-to-end trainable \ac{rag}, ETRAG, which replaces the $k$-nearest neighbor method with differentiable operations and uses obtained instances as soft prompts. ETRAG enables an entire model, including the retriever, to fine-tune for the relation extraction task.
    \item We confirm that ETRAG consistently improves extraction performance for the TACRED, a benchmark for relation extraction, and is particularly effective in situations where training data is limited.
    \item Our analysis reveals that ETRAG can select instances strongly related to the target task. For the relation extraction, instances with the same relationship labels as the extraction target and instances containing the same surface entities account for more than 70\% of the instances.
\end{itemize}

The remainder of this paper is organized as follows: \cref{sec:related-work} presents related work, further elaborating on the evolution and current state of \ac{plm} (\cref{ssec:plm}), instance-based methods (\cref{ssec:instance-base}), and relation extraction (\cref{ssec:re}). \cref{ssec:re} contains notations and explanations used in subsequent sections. \cref{sec:methods} explains our methodology by describing the proposed method ETRAG with differentiable \ac{knn} (\cref{ssec:diff-knn}) and neural prompting (\cref{ssec:neural-prompting-retriever}), and the training methods of ETRAG (\cref{ssec:train}).
The evaluations are performed in \cref{sec:exp} for the extraction performance and \cref{sec:retriever-analysis} for the retriever analysis. Finally, \cref{sec:conclusion} concludes this study and describes future directions.


\section{Related Work}
\label{sec:related-work}

\subsection{Pretrained Language Models}
\label{ssec:plm}
\acp{plm} are large-scale neural network models trained on a large corpus with \ac{nlp} tasks and are usually fine-tuned when applied to various tasks.
Pretraining tasks are extensive such as language modeling (LM)~\cite{radford-2019-gpt-2}, bidirectional language modeling (biLM)~\cite{peters-etal-2018-elmo}, sequence-to-sequence language modeling (Seq2Seq LM)~\cite{raffael-et-al-2020-t5}, and masked language modeling (MLM)~\cite{devlin-etal-2019-bert}. \acp{plm} employ appropriate neural network structures such as ELMo~\cite{peters-etal-2018-elmo} for the bidirectional language modeling task with LSTMs~\cite{hochreiter-1997-lstm}, and Transformer~\cite{vaswani-2017-atttentionisallyouneed-transformer} is mainly used recent systems including BERT~\cite{devlin-etal-2019-bert} for masked language modeling, T5~\cite{raffael-et-al-2020-t5} for sequence-to-sequence language modeling, and GPT-2~\cite{radford-2019-gpt-2} for language modeling. \acp{plm} are extended to a larger scale and use a larger corpus such as Flan-T5~\cite{chung-2022-flan-t5} in Seq2Seq LM and GPT-3 in LM~\cite{brown-etal-2020-gpt3}. 

\acp{plm} are stochastic models to estimate the probability of word sequences based on the pretraining tasks; collectively, these are called text generation models $G$. Since LM is a task to predict the subsequent text from a given input, the model trained on LM can estimate the probability of input text sequence, formulated as $G(x)$ with input $x$. On the other hand, the model trained on Seq2Seq LM can assign a probability to input-output pairs used in pretraining, formulated as $G(y|x)$ with input $x$ and output $y$.

\acp{plm} are computationally expensive to fine-tune the entire parameters when adapted to the target task. Thus, they are tuned by controlling the generated text with text prompts or by parameter-efficient tuning. \ac{icl} performs the target task learned from the context of tuned text prompts~\cite{brown-etal-2020-gpt3}. However, the performance on the target task is often lower than that of a model tuned specifically to the target task, and the prompt tuning process depends on the expert. 
Instead of the text prompt, prompt tuning~\cite{lester-etal-2021-prompt-tuning} or prefix-tuning~\cite{li-liang-2021-prefix-tuning} trains soft prompts, which are trainable vectors inserted to embedded text sequence. Neural prompting~\cite{tian-2023-graph-neural-prompting-arxiv} has extended soft prompts to create them by encoding external information rather than pre-prepared embeddings. 
Since such prompt-based tuning can tune the context but not the model parameters, lightweight model tuning can also be used~\cite{houlsby-2019-adapter,hu-2022-lora}. Low-Rank Adaptation (LoRA)~\cite{hu-2022-lora} inserts low-rank parameters into the base model to learn the difference between the target task. We utilize the LoRA in our experiments to update \ac{plm} and the neural prompting in our method to introduce instances.

\subsection{Retrieval-based Methods}
\label{ssec:instance-base}
Retrieval-based methods, as typified by a nearest neighbor method~\cite{kaplan-1958-original-nearest-neighbor}, have been used for various \ac{nlp} tasks, such as Part-Of-Speech (POS) tagging~\cite{sogaard-2011-semi-nearest-neighbor-pos}, named entity recognition~\cite{yang-katiyar-2020-simple-nearest-neighbor-ner}, dependency parsing~\cite{ouchi-etal-2021-instance-based-dependency}, and relation extraction~\cite{mani-2003-knn-information-extraction}. The methods are used to mitigate a training data scarcity situation. This paper replaces the indifferentiable operations in the \ac{knn} algorithm with differentiable ones in order to relax to a soft operation.

Recently, \acp{plm} are often used following the pretraining task to use instances by \ac{icl}, which predicts the following context from the given prompt. Since \ac{icl} is characterized by the prompt design, the instance selection plays a vital role. Since fixed prompts are usually used, the instances are also typically fixed. However, the demand to select the most appropriate instances for input has led to use \ac{rag}~\cite{patric-2020-rag-retrieval-augumented-generation}, which trains a retriever in advance and uses the instances obtained by the retriever. Since the instance selection operation is not differentiable because of sampling, the general retriever is trained separately to the target task~\cite{wan-etal-2023-gpt-re}. This study tackles this separate training of the target task model and the retriever so that it can be trained end-to-end with the target task.

\subsection{Relation Extraction}
\label{ssec:re}
The fundamental relation extraction task is sentence-level relation extraction~\cite{walker2005ace,hendrickx-etal-2010-semeval}, where only the entity pairs in a single sentence are targets for extraction. Since sentence-level relation extraction ignores relations across sentences, it is extended to document-level relation extraction~\cite{sahu-etal-2019-inter,christopoulou-etal-2019-connecting}, which also extracts these relations. Since this paper focuses on a retriever that retrieves instances tied to a single relation for simplicity, we develop an extractor primarily on sentence-level relation extraction.

Historically, relation extractors have evolved from traditional approaches to modern deep learning techniques. Initially, rule-based systems relied on hand-crafted rules to identify relations~\cite{riloff-1993-rule-based-relation-extraction}. Kernel-based models and feature-based approaches later emerged, offering more flexibility and better handling linguistic variations~\cite{zelenko2003kernel,miwa-sasaki-2014-modeling-joint-relation-close-first}. Deep learning revolutionized the field, introducing models that could learn complex patterns and relationships directly from data~\cite{zeng-etal-2014-first-deep-learning-relation-extraction}.

Given input sentence $x$ with head entity $h$ and tail entity $t$ of a target entity pair and relation $r \in R$ between them, a standard relation extractor in \cref{eq:relation-extraction-classification} formulates a classification task using a stochastic model $P$ which is constructed with deep learning.
\begin{equation}
    \label{eq:relation-extraction-classification}
    \hat{r} = \argmax_{r \in R} P (r|x,h,t)
\end{equation}

\subsubsection{Relation Extraction with \aclp{plm}}

Most recent relation extractors employ \acp{plm} for modeling $P$ as feature extractors to prepare the feature vector for direct classification~\cite{zhou-2021-atlop-relation-cdr-docred} or prepare the features for the following relation extraction-specific model~\cite{fu-etal-2019-graphrel}.
On the other hand, \acp{plm} are also used as the text generation model with prompt engineering~\cite{wan-etal-2023-gpt-re} or fine-tuning~\cite{wadhwa-etal-2023-revisiting-llm-relation-extraction}. For example, SuRE (Summarization as Relation Extraction)~\cite{lu-etal-2022-summarization-sure} extracts the relation with summarization via \ac{plm} and mapping the summarized text output to relations. SuRE measures the probability of a pair of input text prompts and verbalizes relations in a summary form to predict the relation with the highest probability.

Typical text generation-based extractors prepare input and output templates with placeholders to fill with information to ask about relations. Let an input template be $\mathrm{Template}_{\mathrm{in}}(x,h,t)$, an output template be $\mathrm{Template}_{\mathrm{out}}(x,h,t,r)$. \cref{eq:template-in} and \cref{eq:template-out} show examples of them.
\begin{align}
    &\begin{multlined}
        \label{eq:template-in}
        \mathrm{Template}_{\mathrm{in}} (x,h,t) = \text{``The head entity is } h \text{ . } \\ \text{The tail entity is } t \text{ . } x \text{''} 
    \end{multlined}\\
    &\begin{multlined}
        \label{eq:template-out}
        \mathrm{Template}_{\mathrm{out}} ( x , h , t , r = \text{no\_relation} ) = \\ \text{``} h \text{ has no known relations to } t \text{ .''} 
    \end{multlined}
\end{align}
Let a text generation model that computes the probability of the output template conditioned by the input sequence be $G(\mathrm{Template}_{\mathrm{out}} \mid \mathrm{Template}_{\mathrm{in}})$, \cref{eq:llm-re} represents the text generation-based relation extractor.
\begin{equation}
    \label{eq:llm-re}
    \hat{r} = \argmax_{r \in R} G(\mathrm{Template}_{\mathrm{out}}(x,h,t,r) \mid \mathrm{Template}_{\mathrm{in}}(x,h,t))
\end{equation}

SuRE is a typical method for relation extraction in this form, which improves the efficiency of text-generation-based relation extraction according to \cref{eq:llm-re}. Calculating probabilities for all relation labels in a straightforward manner is expensive due to the large number of calculations required by \ac{plm}. Therefore, SuRE prepares a trie tree of templates for all relation labels and searches for the template with the highest probability by beam search on the trie, thereby realizing the prediction of relations by text generation with a small number of calculations. Note that the model training is the same as that of normal text generation models.

\subsubsection{Relation Extractor Using Instances}
Despite recent advancements in deep learning models, relation extraction remains challenging, primarily due to the scarcity of annotated data. Deep learning models, in particular, require large amounts of labeled data for training~\cite{miwa-bansal-2016-end}. However, manually annotating data is expensive and time-consuming, making it a significant bottleneck in developing effective relation extraction systems~\cite{han-etal-2018-fewrel}.
This environment makes it difficult to classify with high performance for complex relations types.
In this context, the relation extractor via the nearest neighbor approach~\cite{wan-etal-2022-rescue-nearest-neighbor-relation-extraction}, where relation extractors leverage similar instances during inference, has shown promise in efficiently utilizing limited data. The goal of this study is to reveal that using instances, or specific instances of relations within texts, is crucial in relation extraction. By leveraging these instances, models can better generalize from limited instances and improve their ability to extract and classify relations in varied contexts accurately.

The relation extractors with the text generation model also use instances by introducing selected instances as text into the prompt~\cite{wan-etal-2023-gpt-re,wadhwa-etal-2023-revisiting-llm-relation-extraction}. Such methods extend the templates to accept selected instances so that the text generation model can handle instances by preparing placeholders for them. When using a retriever~\cite{wan-etal-2023-gpt-re}, instances are prepared according to the input, and when not using a retriever~\cite{wadhwa-etal-2023-revisiting-llm-relation-extraction}, instances are prepared in advance manually.

The instances enhanced text generation model involves two processes: the retrieved instances into prompts compatible with the text generation model. In the traditional \ac{rag} framework~\cite{patric-2020-rag-retrieval-augumented-generation}, the retriever identifies neighboring instances and directly utilizes the text of these instances as prompts. Conversely, in neural prompting~\cite{tian-2023-graph-neural-prompting-arxiv}, instances are selected based on string matching and processed through a neural network to use the instances as soft prompts, offering a more nuanced and adaptable approach to instance utilization.
Let an external database for a reference, which includes the elements of a text and relation information $(x,h,t,r)$, be $D$, a set of selected instances be $D' \subseteq D$, the template that accepts instances as input be $\mathrm{Template}_{\mathrm{in}}(x,h,t,D')$, and the retrieval process to obtain $D'$ from $D$ using input information be $D' = \mathrm{Retrieve}(x,h,t,D)$, the text generation-based relation extractor with instances are shown in \cref{eq:retrieval-augumented-re}.
\begin{multline}
    \label{eq:retrieval-augumented-re}
    \hat{r} = \argmax_{r \in R} G(\mathrm{Template}_{\mathrm{out}}(x,h,t,r) \mid \\ \mathrm{Template}_{\mathrm{in}}(x,h,t, D'))
\end{multline}

In the retriever that searches nearest instances in the embedding space, each instance $d$ in the database $D$ is converted into $L$ embeddings $E_{d}$:
\begin{align}
E_{d}= [E_{1,d}, E_{2,d}, \dots, E_{L,d}]^{\top} \in \mathbbm{R}^{L \times N}. 
\end{align}

The retriever aims to select instances that are near the input embedding $E_{\mathrm{in}} \in \mathbbm{R}^{L \times N}$, which is embedded similarly to $E_{d}$ using $x$, $h$, and $t$. The method to embed instances and the input is usually engineered to suit the purpose; for example, modern applications~\cite{wu-etal-2023-openicl} often use \acp{plm} such as BERT~\cite{devlin-etal-2019-bert}. Our method also uses \ac{plm} to embed instances.
Nearest instance search requires the distance between input and each instance calculated with a function to compute distance as $\mathrm{Dist}(E_{\mathrm{in}}, E_{d})$.
Since the \ac{knn} retriever's objective is to emit instances within a distance of up to $k$-th, a set of target instances $D'$ becomes \cref{eq:target-instances}, where $\argtopk$ returns a set of top-$k$ indices:
\begin{equation}
    \label{eq:target-instances}
    D' = \left\{D_i \mid i \in \argtopk_j - \mathrm{Dist}(E_{\mathrm{in}}, E_j) \right\}
\end{equation}
After the retrieval, the instances are converted into a format that can be input into the model in the embedding process, e.g., the traditional method writes down into text prompts for \ac{plm}~\cite{patric-2020-rag-retrieval-augumented-generation}.

\section{Methodology}
\label{sec:methods}

\begin{figure*}
    \centering
    \includegraphics[width=0.85\linewidth]{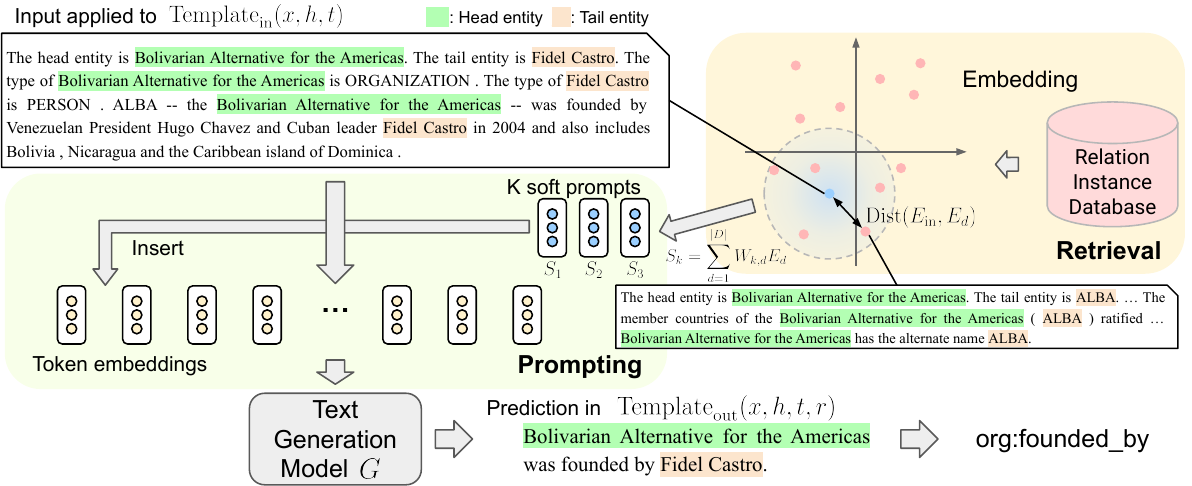}
    \caption{An overview of ETRAG}
    \label{fig:method}
\end{figure*}

We propose ETRAG, which enables end-to-end training of text generation models with \ac{rag}. In order to fit a model pretrained for a general task to the target task, the parameters need fine-tuning to the objective of the target task. However, the retriever part of general \ac{rag} cannot be trained end-to-end, and the instances selection cannot be optimized when training the model on the target task. The entire model must consist of differentiable operations to compute gradient when training a deep learning model end-to-end. However, two indifferentiable operations prevent end-to-end training: selecting instances by retriever and making the selected instances into a text prompt.

To overcome the indifferentiable processes, ETRAG replaces the retriever's process of selecting instances with a soft $k$NN and introduces instances as soft prompts as shown in \cref{fig:method}.
For the retriever selection, we employ a soft $k$NN, which selects instances softly inspired by neural nearest neighbor networks~\cite{tobias-2018-neural-nearest-neighbors-networks}.
When introducing selected instances, ETRAG injects the instances to input as soft prompts, which concatenates instance embeddings into embedded input tokens rather than text prompts, like neural prompting~\cite{tian-2023-graph-neural-prompting-arxiv}.

To extract relations by a text-generating model, we employ SuRE~\cite{lu-etal-2022-summarization-sure} as the base relation extractor. SuRE is a method of inference by a text-generating model, and since the model is a regular text-generation model, we propose a method for introducing instances for the text-generation model.

We will explain the method in the following sections: ETRAG consisting of differentiable $k$-nearest instance selection (\cref{ssec:diff-knn}) and integration of the instances drawn by it (\cref{ssec:neural-prompting-retriever}). The end-to-end training and training techniques for ETRAG are shown in \cref{ssec:train}. 

\subsection{Differentiable $k$-Nearest Instance Selection}
\label{ssec:diff-knn}
The differentiable $k$-nearest instance selection is achieved by weighted selection over multiple instances. In contrast, the standard $k$-nearest instance selection is indifferentiable because instances are selected with sampling operation as shown in \cref{eq:target-instances}. Therefore, in this section, we first describe the embedding of instances created by weighted sums of the instances and then explain the creation of weights.

We formulate the differentiable instance selection by creating a weighted sum of the instance embeddings, paying attention to the important instances. Let weights be $W \in \mathbbm{R}^{K \times |D|}$, where $W_{k,d}$ means the weight of $d$-th data for $k$-th selection, $k$-th selected instance embedding $S_k \in \mathbbm{R}^{L \times N}$ is defined as $S_k = \sum_{d=1}^{|D|} W_{k,d} E_d$.

We then explain how to compute the weights in a differentiable manner.
The simplest solution to sample the nearest instance is applying softmax to the distance between instances such as Gumbel softmax~\cite{jang-2017-categorical-gumbel-softmax}. However, this approach is possible to select only the nearest neighbor instance, not the second, third, and subsequent instances. In order to realize subsequent selection, instances with heavy weights in previous steps are penalized, and the weights in subsequent steps are computed based on them so that the instances once heavily weighted process almost exclusively, as described in \cref{eq:etrask}.
\begin{align}
    \label{eq:etrask}
    W_{k,d} = 
    \begin{cases}  \frac{\exp{( 1 - \mathrm{Dist}(E_{\mathrm{in}}, E_d))}}{\sum_{d'=1}^{|D|} \exp{( 1 - \mathrm{Dist}(E_{\mathrm{in}}, E_{d'}))}} \qquad \text{if $k = 1$} \\
     {\scriptsize \begin{multlined}  \frac{\exp{( 1 - \mathrm{Dist}(E_{\mathrm{in}}, E_d)} + \sum_{l=1}^{k-1} \log (1-W_{l,d})) }{\sum_{d'=1}^{|D|} \exp{( 1 - \mathrm{Dist}(E_{\mathrm{in}}, E_{d'})} + \sum_{l=1}^{k-1} \log (1-W_{l,d'})) } \\ \text{otherwise} \end{multlined}}
    \end{cases}
\end{align}
The weights $W_k$ become a nearly one-hot vector since the softmax function computes the selection weight.
We assume the distance $\mathrm{Dist}(E_{\mathrm{in}}, E_{d})$ is an average of the cosine distance between $E_{\mathrm{in}}$ and $E_{d}$ as defined in \cref{eq:distance-def}, where $\mathrm{Dist}$ is bounded from 0 to 1 ($0 \leq \mathrm{Dist} \leq 1$).
\begin{equation}
    \label{eq:distance-def}
    \mathrm{Dist}(E_{\mathrm{in}}, E_{d}) = \frac{1}{2L} \sum_{l=1}^L 1 - \frac{E_{\mathrm{in},l}^\top E_{d,l}}{|E_{\mathrm{in},l}||E_{d,l}|}
\end{equation}

The retrieval for relation extraction requires the preparation of embeddings $E$. We use representations of head and tail entities and relation labels, which are considered helpful for relation extraction~\cite{verga-etal-2018-bran}. These representations are obtained by writing down the relation instances in text using templates, embedding them using another \ac{plm}, and extracting the representations of the text corresponding to the head entity, tail entity, and relation labels. Specifically, we use an average of embeddings in the span of entities or relations, where $E_{\cdot, 1}$ and $E_{\cdot, 2}$ are the embeddings of head and tail entity and $E_{\cdot , 3}$ is the embedding of relation.

These processes provide $K$ weights for each instance and $K$ selected instances. 
When using this $k$-nearest instance selection, the neural model constructed using the weights to select instances can select instances without losing trainability and obtain their embeddings.

\subsection{Neural Prompting with Trainable Instance Selection}
\label{ssec:neural-prompting-retriever}
We propose a neural prompting that creates soft prompts for the text generation model from the selection weights via differentiable $k$-nearest instance selection while the general retriever creates prompts as text. We compose soft prompts from instances softly selected by weighted summing of the embeddings over the instances using the selection weights in \cref{ssec:diff-knn} instead of text prompts.
The retriever becomes differentiable by composing with this process.

For the detailed procedure, the $k$-th selected instance embedding $S_k$ is computed in the retrieval process. Now, we need to prepare soft prompts from selected instances by reshaping selected instance embeddings $S$ to the shape for a soft prompt $P \in \mathbbm{R}^{KL \times N}$ by stacking them as $P = [S_{1,1}, S_{1,2}, \dots S_{1,L}, S_{2,1}, \dots S_{K,L}]$.
Connecting the prompt and a text generation model involves joining the prompt to the input sequence embeddings. In this case, the length of the prompt $KL$ is added to the length of the input series $|x|$, resulting in a new series with the length of $|x| + KL$.

\subsection{Training}
\label{ssec:train}

The training of the retriever proposed in this paper is simply a matter of optimizing the ETRAG model with the objective function of the target task. Since the retriever consists entirely of differentiable operations in ETRAG, the model with the retriever is end-to-end trainable. Our method innovatively transforms the retriever into an end-to-end trainable model, enhancing its applicability to relation extractors.

The primary challenge in training the retriever is its computational intensity, which requires calculating distances for all instances in $D$. To mitigate this during training, we employ a strategy of random sampling a subset of instances. This approach significantly reduces the computational burden while maintaining the retriever's efficacy.

Since the base model, SuRE, is subsequently connected to the retriever, a stable training method for the base model is needed. The training process of the text generation model easily affects the retriever's performance. Therefore, we employ a warm-up step in which the retriever is trained in advance. The base model is frozen, and the retriever is updated during the warm-up steps. All the parameters are updated after the warm-up step.

\section{Evaluation of Relation Extraction Performance}
\label{sec:exp}
This section evaluates the relation extraction performance and compares the ETRAG integrated model to existing relation extraction methods. \cref{ssec:exp-setting} presents the settings of subsequent experiments about datasets, baseline methods, model settings, and training parameters. Based on the settings, \cref{ssec:result} shows the performance evaluations of our proposal by comparing it with other methods. \cref{ssec:ablation} is the ablation study to show the elements that affect performance.

\subsection{Experimental Settings}
\label{ssec:exp-setting}

\begin{table}[t]
    \centering
    \caption{Statistics of the TACRED Dataset with Different Proportion.}
    \small
    \begin{tabular}{rrrr}
    \hline
        Proportion & Train & Dev. & Test \\
         \hline
        100\% & 68,125 &  22,631 & \multirow{4}{*}{15,509} \\
        10\% & 6,815 & 2,265 \\
        5\% & 3,407 & 1,133 \\
       1\% & 682 & 227  \\
        \hline
    \end{tabular}
    \label{tab:dataset-stat}
\end{table}

\begin{table*}[t]
    \centering
    \caption{Statistics of Relation Labels for Each Split and Proportion of TACRED dataset.}
    \begin{tabular}{l|rr|rr|rr|rr|r}
        \hline
        & \multicolumn{2}{|c|}{100\%} & \multicolumn{2}{c|}{10\%} & \multicolumn{2}{c|}{5\%} & \multicolumn{2}{c|}{1\%} \\
        Relation Label & Train & Dev. & Train & Dev. & Train & Dev. & Train & Dev. & Test  \\
        \hline
        no\_relation & 55,112 & 17,195 & 5,513 & 1,721 & 2,756 & 861 & 552 & 173 & 12,184\\
        per:title & 2,443 & 919 & 244 & 92 & 122 & 46 & 24 & 9 & 500\\
        org:top\_members/employees & 1,890 & 534 & 190 & 53 & 95 & 26 & 19 & 5 & 346\\
        per:employee\_of & 1,524 & 375 & 152 & 38 & 76 & 19 & 15 & 4 & 264\\
        org:alternate\_names & 808 & 338 & 80 & 34 & 40 & 17 & 8 & 3 & 213\\
        org:country\_of\_headquarters & 468 & 177 & 47 & 18 & 24 & 9 & 5 & 2 & 108\\
        per:countries\_of\_residence & 445 & 226 & 44 & 22 & 22 & 11 & 4 & 2 & 148\\
        per:age & 390 & 243 & 40 & 24 & 20 & 12 & 4 & 2 & 200\\
        org:city\_of\_headquarters & 382 & 109 & 38 & 10 & 19 & 5 & 4 & 1 & 82\\
        per:cities\_of\_residence & 374 & 179 & 38 & 18 & 19 & 9 & 4 & 2 & 189\\
        per:stateorprovinces\_of\_residence & 331 & 72 & 33 & 8 & 16 & 4 & 3 & 1 & 81\\
        per:origin & 325 & 210 & 32 & 21 & 16 & 11 & 3 & 2 & 132\\
        org:subsidiaries & 296 & 113 & 30 & 12 & 15 & 6 & 3 & 1 & 44\\
        org:parents & 286 & 96 & 28 & 10 & 14 & 5 & 3 & 1 & 62\\
        per:spouse & 258 & 159 & 26 & 16 & 13 & 8 & 3 & 2 & 66\\
        org:stateorprovince\_of\_headquarters & 229 & 70 & 23 & 7 & 11 & 4 & 2 & 1 & 51\\
        per:children & 211 & 99 & 21 & 10 & 10 & 5 & 2 & 1 & 37\\
        per:other\_family & 179 & 80 & 18 & 8 & 9 & 4 & 2 & 1 & 60\\
        org:members & 170 & 85 & 17 & 8 & 9 & 4 & 2 & 1 & 31\\
        per:siblings & 165 & 30 & 17 & 3 & 9 & 1 & 2 & 0 & 55\\
        per:parents & 152 & 56 & 16 & 6 & 8 & 3 & 2 & 1 & 88\\
        per:schools\_attended & 149 & 50 & 14 & 5 & 7 & 3 & 1 & 1 & 30\\
        per:date\_of\_death & 134 & 206 & 13 & 20 & 6 & 10 & 1 & 2 & 54\\
        org:founded\_by & 124 & 76 & 12 & 8 & 6 & 4 & 1 & 1 & 68\\
        org:member\_of & 122 & 31 & 12 & 3 & 6 & 1 & 1 & 0 & 18\\
        per:cause\_of\_death & 117 & 168 & 12 & 17 & 6 & 9 & 1 & 2 & 52\\
        org:website & 111 & 86 & 11 & 9 & 5 & 5 & 1 & 1 & 26\\
        org:political/religious\_affiliation & 105 & 10 & 10 & 1 & 5 & 0 & 1 & 0 & 10\\
        per:alternate\_names & 104 & 38 & 10 & 4 & 5 & 2 & 1 & 0 & 11\\
        org:founded & 91 & 38 & 10 & 4 & 5 & 2 & 1 & 0 & 37\\
        per:city\_of\_death & 81 & 118 & 8 & 12 & 4 & 6 & 1 & 1 & 28\\
        org:shareholders & 76 & 55 & 8 & 6 & 4 & 3 & 1 & 1 & 13\\
        org:number\_of\_employees/members & 75 & 27 & 8 & 2 & 4 & 1 & 1 & 0 & 19\\
        per:charges & 72 & 105 & 8 & 10 & 4 & 5 & 1 & 1 & 103\\
        per:city\_of\_birth & 65 & 33 & 7 & 3 & 4 & 1 & 1 & 0 & 5\\
        per:date\_of\_birth & 63 & 31 & 7 & 3 & 4 & 1 & 1 & 0 & 9\\
        per:religion & 53 & 53 & 6 & 6 & 3 & 3 & 1 & 1 & 47\\
        per:stateorprovince\_of\_death & 49 & 41 & 4 & 4 & 2 & 2 & 0 & 0 & 14\\
        per:stateorprovince\_of\_birth & 38 & 26 & 4 & 2 & 2 & 1 & 0 & 0 & 8\\
        per:country\_of\_birth & 28 & 20 & 2 & 2 & 1 & 1 & 0 & 0 & 5\\
        org:dissolved & 23 & 8 & 2 & 0 & 1 & 0 & 0 & 0 & 2\\
        per:country\_of\_death & 6 & 46 & 0 & 5 & 0 & 3 & 0 & 1 & 9\\
        \hline
    \end{tabular}
    \label{tab:relation-label-stats}
\end{table*}

To assess the effectiveness of our relation extraction method, we experiment on the TACRED dataset~\cite{zhang-etal-2017-position-relation-sentence-tacred}, a standard benchmark in the sentence-level relation extraction, where each instance has an entity pair within a sentence and a gold relation between the pair. TACRED is the only dataset for which templates are available, as it is the dataset of the target evaluated by the base model of SuRE. To understand the impact of training data size, we experimented with reduced training and development data scenarios for TACRED: 100\%, 10\%, 5\%, and 1\% of the entire dataset, following existing study~\cite{sainz-etal-2021-label-verbalization-nli-deberta}. The statistics for each scenario are shown in \cref{tab:dataset-stat,tab:relation-label-stats}. 
We evaluated the mean value of the micro-averaged F1 score for three runs, treating the \texttt{no\_relation} class as a negative example for TACRED following the official evaluation settings.
We calculate the loss to development data for every 100 update steps and report the score obtained when training in early stopping with the early stopping patience of 3 for the TACRED dataset.

We compared our method against state-of-the-art models: SuRE (based on Pegasus-large)~\cite{lu-etal-2022-summarization-sure}, DeepStruct~\cite{wang-etal-2022-deepstruct-plm-relation-extraction}, kNN-RE~\cite{wan-etal-2022-rescue-nearest-neighbor-relation-extraction}, and NLI\_DeBERTa~\cite{sainz-etal-2021-label-verbalization-nli-deberta}. SuRE is, as described in \cref{ssec:re}, the model extracting relation with the text generation model by formulating the relation extraction task to the text summarization task using Seq2Seq LM. DeepStruct is a \ac{plm} trained for structured prediction tasks, which include relation extraction. kNN-RE performs \ac{knn} algorithm on the \ac{plm} embedding space of relation instances.
NLI\_DeBERTa extracts relations with a natural language inference task, which recognizes fact inclusion in a hypothesis, by identifying the implication of verbalized relations in a target text.

\begin{figure}[t]
    \centering
        \begin{minipage}[t]{\linewidth}
                \small
                \centering
        \fbox{\begin{minipage}{0.96\linewidth}
        The head entity is \underline{\$\{head entity\}}. The tail entity is \underline{\$\{tail entity\}}. The type of \$\{head entity\} is \$\{label of head entity\} . The type of \$\{tail entity\} is \$\{label of tail entity\} . \$\{input sentence\} .
            \end{minipage}}
        \end{minipage}
    \caption{The input template of SuRE. \$\{$\cdot$\} are placeholders to replace with input.}
    \label{fig:template}
\end{figure}

\begin{table*}[t]
    \centering
    \caption{The output template of SuRE. \$\{$\cdot$\} are placeholder replaced with input.}
    \begin{tabular}{ll}
        \hline
        Relation Label & Template \\
        \hline
        no\_relation & \$\{subj\} has no known relations to \$\{obj\} \\
        per:title & \$\{subj\} is a \$\{obj\} \\
        org:top\_members/employees & \$\{subj\} has the high level member \$\{obj\} \\
        per:employee\_of & \$\{subj\} is the employee of \$\{obj\} \\
        org:alternate\_names & \$\{subj\} is also known as \$\{obj\} \\
        org:country\_of\_headquarters & \$\{subj\} has a headquarter in the country \$\{obj\} \\
        per:countries\_of\_residence & \$\{subj\} lives in the country \$\{obj\} \\
        per:age & \$\{subj\} has the age \$\{obj\} \\
        org:city\_of\_headquarters & \$\{subj\} has a headquarter in the city \$\{obj\} \\
        per:cities\_of\_residence & \$\{subj\} lives in the city \$\{obj\} \\
        per:stateorprovinces\_of\_residence & \$\{subj\} lives in the state or province \$\{obj\} \\
        per:origin & \$\{subj\} has the nationality \$\{obj\} \\
        org:subsidiaries & \$\{subj\} owns \$\{obj\} \\
        org:parents & \$\{subj\} has the parent company \$\{obj\} \\
        per:spouse & \$\{subj\} is the spouse of \$\{obj\} \\
        org:stateorprovince\_of\_headquarters & \$\{subj\} has a headquarter in the state or province \$\{obj\} \\
        per:children & \$\{subj\} is the parent of \$\{obj\} \\
        per:other\_family & \$\{subj\} is the other family member of \$\{obj\} \\
        org:members & \$\{subj\} has the member \$\{obj\} \\
        per:siblings & \$\{subj\} is the siblings of \$\{obj\} \\
        per:parents & \$\{subj\} has the parent \$\{obj\} \\
        per:schools\_attended & \$\{subj\} studied in \$\{obj\} \\
        per:date\_of\_death & \$\{subj\} died in the date \$\{obj\} \\
        org:founded\_by & \$\{subj\} was founded by \$\{obj\} \\
        org:member\_of & \$\{subj\} is the member of \$\{obj\} \\
        per:cause\_of\_death & \$\{subj\} died because of \$\{obj\} \\
        org:website & \$\{subj\} has the website \$\{obj\} \\
        org:political/religious\_affiliation & \$\{subj\} has political affiliation with \$\{obj\} \\
        per:alternate\_names & \$\{subj\} has the alternate name \$\{obj\} \\
        org:founded & \$\{subj\} was founded in \$\{obj\} \\
        per:city\_of\_death & \$\{subj\} died in the city \$\{obj\} \\
        org:shareholders & \$\{subj\} has shares hold in \$\{obj\} \\
        org:number\_of\_employees/members & \$\{subj\} has the number of employees \$\{obj\} \\
        per:charges & \$\{subj\} is convicted of \$\{obj\} \\
        per:city\_of\_birth & \$\{subj\} was born in the city \$\{obj\} \\
        per:date\_of\_birth & \$\{subj\} has birthday on \$\{obj\} \\
        per:religion & \$\{subj\} has the religion \$\{obj\} \\
        per:stateorprovince\_of\_death & \$\{subj\} died in the state or province \$\{obj\} \\
        per:stateorprovince\_of\_birth & \$\{subj\} was born in the state or province \$\{obj\} \\
        per:country\_of\_birth & \$\{subj\} was born in the country \$\{obj\} \\
        org:dissolved & \$\{subj\} dissolved in \$\{obj\} \\
        per:country\_of\_death & \$\{subj\} died in the country \$\{obj\} \\
        \hline
    \end{tabular}
    \label{tab:sure-temp}
\end{table*}

We employed SuRE~\cite{lu-etal-2022-summarization-sure} as a generation-based relation extraction model. We introduced the ETRAG to SuRE by adding soft prompts into the input sequence between the BOS token and the following prompt. 
The hyperparameters of the SuRE were the same as those of the original research. The templates were the same as in the original paper: the templates for the SuRE input were in the form of \cref{fig:template}, from which the summary templates defined for each relation label are predicted as in \cref{tab:sure-temp}. 
The beam search width, a parameter used in SuRE classification, was set to 4, the same value as in the original paper of SuRE.

We use the Flan-T5 large model~\cite{chung-2022-flan-t5} in the SuRE framework with and without the addition of ETRAG because the tuning before experiments showed the training of Pegasus-large~\cite{zhang-2020-pegasus} based SuRE with ETRAG was unstable. When we introduced ETRAG into the Pegasus-based model, the model predicted only \texttt{no\_relation} for the same setting. Since our method used two \acp{plm}, the base relation extraction model and the embedding model for the retriever, the larger models were unacceptable for our computational resources. We conducted trials introducing 10-neighbor instances as prompts, i.e., $K = 10$, pretraining the retriever for 300 steps before end-to-end training as the warm-up step described in \cref{ssec:train}. Due to computational constraints, the database $D$ was constructed from randomly sampled 5,000 instances in the training dataset if training data has more than 5,000 instances. At the training time, 32 instances are randomly sampled as the subset of $D$ before retrieval. 
For the embeddings $E$, we used the entity and relation representations of \ac{plm} by averaging their spans, where the \ac{plm} input was created by concatenating the template in \cref{fig:template} with the relation template. 
The entity spans were underlined parts of \cref{fig:template}. The relation span was the relation template part when the database was embedded and the EOS token when the prediction target was embedded. We applied LoRA~\cite{hu-2022-lora} to Flan-T5 with rank $r=32$ and dropout rate 0.1 to all kinds of layers of Transformer. The LoRA is a method for efficient fine-tuning, and we employed it because our pilot experiments showed full tuning and LoRA tuning performances are not significantly different. The dropout rate was set to 0.1. AdamW was used to optimize the models, with a learning rate of $5\times10^{-4}$ for Flan-T5 and $1\times10^{-3}$ for the other parameters, and weights of $5\times10 ^{-6}$ for the bias and layer normalization parameters. The batch size was set to 64. The parameters used for evaluation were those used when early stopping completed training with patience set to 5. A single NVIDIA A100 was used for each experiment.

\subsection{Extraction Performance Comparison}
\label{ssec:result}

\begin{table}[t]
    \centering
    \caption{Comparison of Relation Extraction Performance [\%]}
    \begin{tabular}{l|rrrr}
        \hline
         & 100\% & 10\% & 5\% & 1\%  \\
         \hline
         DeepStruct~\cite{wang-etal-2022-deepstruct-plm-relation-extraction} &  76.8 & -- & -- & -- \\
         SuRE (Pegasus)~\cite{lu-etal-2022-summarization-sure} & 75.1 & 70.7 & 64.9 & 52.0 \\
         NLI\_DeBERTa~\cite{sainz-etal-2021-label-verbalization-nli-deberta} & 73.9 & 67.9 &69.0 & 63.0\\
         kNN-RE~\cite{wan-etal-2022-rescue-nearest-neighbor-relation-extraction} & 70.6 & -- & -- & --\\
        \hline
        SuRE (Flan-T5) & 71.4 \pmm 1.6 & 68.5 \pmm 1.5 & 65.0 \pmm 3.1 & 53.5 \pmm 1.4  \\
        + ETRAG & 73.3 \pmm 0.5 & 71.5 \pmm 0.5 & 68.3 \pmm 2.0 & 54.6 \pmm 1.1 \\
        \hline
    \end{tabular}
    \label{tab:main-result}
\end{table}

The results in \cref{tab:main-result} indicate that ETRAG consistently improved performance from the model without ETRAG for the TACRED dataset. The results confirm that ETRAG can enhance relations extraction by text generation. Additionally, ETRAG outperforms the existing models, SuRE and NLI\_DeBERTa, in scenarios with limited training data (10\%). This is the new state-of-the-art result for the setting of the TACRED dataset under the 10\% setting.

Comparing Pegasus-based and T5-based SuRE, the Pegasus-based SuRE performed better. This is because Pegasus is a model created specifically for the summarization task, which matches SuRE's objective. Even in this situation, the ETRAG boosted the performance of the Flan-T5-based model and achieved the best performance on the 10\% setting.

Compared to another instance-based method, kNN-RE, adding generation-based prediction of relations to neighborhood method-based inference confirms the improved extraction performance. This proves that simple instance utilization is insufficient and that inference capability is essential.

\subsection{Ablation Studies}
\label{ssec:ablation}
This section delves into the factors influencing the extraction performance observed in \cref{ssec:result} and investigates the model's behavior. We conducted ablation studies in the 10\% training instance setting for the TACRED dataset to confirm when our method showed notable improvements.

Our ablation study aimed to pinpoint the elements critical to our method's enhanced performance. We examined various scenarios: employing k-nearest neighbor instances without retriever training (\emphname{No Retriv. Training}), omitting the warm-up phase in Retriever training (\emphname{No Warm-up}), using randomly chosen instances (\emphname{Random}), and utilizing CLS token representations (\emphname{CLS}).\emphname{No Retriv. Training} aims to investigate the usefulness of the end-to-end trainable retriever by using the initial parameter for the retriever and operations of ETRAG. \emphname{No Warm-up} omits the warm-up step but trains the retriever, which checks the stability effect of the warm-up. \emphname{Random} picks up instances randomly and uses soft prompts in the same embedding process as ETRAG, where the experiment checks retrieval process training effectiveness. \emphname{CLS} uses only CLS token representations instead of the relation extraction-specific representation. The other settings of experiments are the same as settings in \cref{ssec:exp-setting} except for the number of runs for evaluation that changed from 3 runs to 1 run.

\begin{table}[t]
    \centering
    \caption{Ablation Study Results [\%]}
    \begin{tabular}{lr}
        \hline
        SuRE (Flan-T5) & 68.5 \\
        + ETRAG & 71.7 \\
        \hline
        \emphname{No Retriv. Training} & 69.2\\
        \emphname{No Warm-up} & 70.7 \\
        \emphname{Random} & 71.0 \\
        \emphname{CLS} & 68.8 \\
        \hline
    \end{tabular}
    \label{tab:ablation}
\end{table}

The results in \cref{tab:ablation} reveal that omitting any of these components results in lower F1 scores, underscoring their collective importance. 
The \emphname{No Retriv. Training} caused a performance loss of 2.2 percentage points, which was not much different from the performance of SuRE without any retrievers. This result indicates that in the relation extraction with text generation model, the retriever needs to be trained for the relation extraction objective to improve performance when relation instances are introduced with the retriever.

The \emphname{No Warm-up} reduced extraction performance by 1.0 percentage points; the decrease was relatively smaller than in the other cases, \emphname{No Retriv. Training} and \emphname{CLS}. This may be due to the lack of treatment for convergence stability, although similar processing and training of the model is carried out.

In the case of \emphname{Random}, where random instances are used without training the retrieval process, the performance drop is relatively small, around 0.7 percentage points. This comparison allows us to evaluate the improvement separately due to the introduction and selection of instances. Compared to the case where no retrieval process is used (i.e., SuRE), introducing randomly selected instances shows a 2.4 percentage point improvement. The results suggest that introducing examples improves performance, and further progress is made when the model selects instances.

For comparison of representations used in ETRAG, the performance with the CLS representation degraded 2.9 percentage points from the one engineered for relation extraction. Additionally, the performance was almost the same as one of SuRE without a retriever. Thus, engineering a representation specializing in the target task is essential to using ETRAG.

Overall, these analyses highlight the contributions of our method, ETRAG, to performance, and we confirmed that all its components are effective. Instance selection and training helped improve relation extraction outcomes.

\section{Retriever Analysis}
\label{sec:retriever-analysis}
\cref{sec:exp} showed the performance improvement by our proposed ETRAG that was integrated into the text generation model (\cref{ssec:result}). The ablation studies in \cref{ssec:ablation} showed the factors contributing to performance. However, it is not yet clear what happened inside ETRAG that led to the improvements. Therefore, we also analyzed the retriever from the perspective of instances.
\cref{ssec:num-instance} confirms the impact of the instances by changing the number of instances created by the retriever and checking their behavior at that time; \cref{ssec:instances} shows what instances were retrieved and used after training by statistically analyzing the instances to analyze the actual retrieved instances directly.

\subsection{Performance Variation with Number of Instances}
\label{ssec:num-instance}
\begin{figure}[t]
    \centering
    \includegraphics[width=0.85\linewidth]{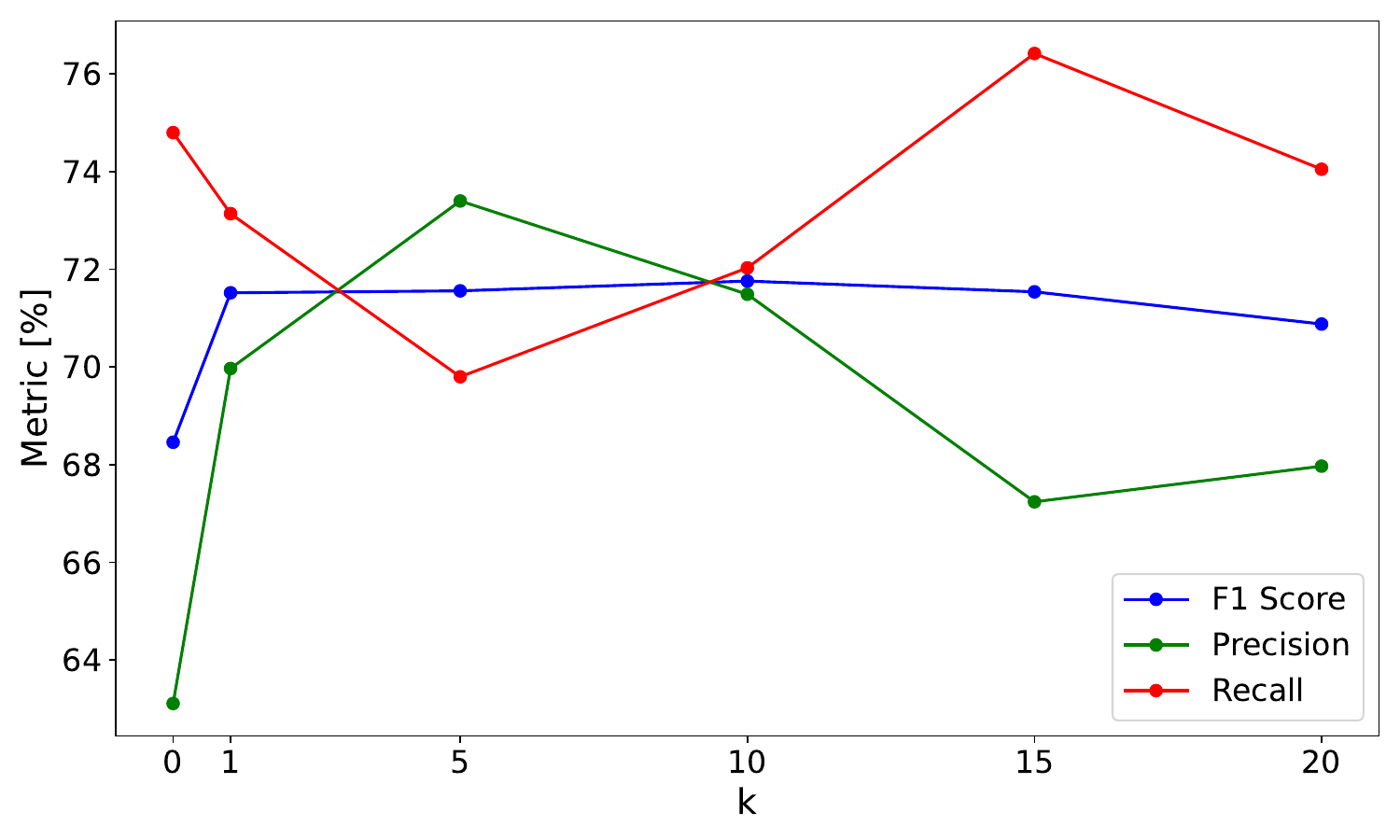}
    \caption{Change in Extraction Performance with the Number of Instances $k$ Used in Prompts}
    \label{fig:num-example-vs-performance}
\end{figure}

We investigated the sensitivity of the number of instances $k$ by varying $k$ from 0 to 20, where 0 instance means ETRAG is not used. The F1, precision, and recall scores versus $k$ are shown in \cref{fig:num-example-vs-performance}. The F1 score reached its maximum at $k=10$ and no significant change in the $1 \leq k \leq 15$ interval. On the other hand, the balance between precision and recall changed significantly. In the $0 \leq k \leq 5$ interval, precision increased, and recall decreased as overall performance improved. Conversely, recall gradually increased, and precision decreased in the interval $k > 5$, except for $k=20$. Since precision and recall were balanced at $k=10$, the F1 score was the largest, defined as the harmonic mean of the precision and recall. 

These characteristics are useful in real-world applications and can be used to make performance trade-offs of precision and recall to suit the situation. For example, applications that require users to retrieve necessary relational information, such as a search system, could use a larger $k$ for coverage.

\subsection{Retrieved Instances}
\label{ssec:instances}
Since the characteristics of retrievers are most evident in retrieved instances, we investigate the retrieved instances in ETRAG. However, because selected instances in ETRAG are a weighted sum of the instance embeddings, obtaining what was chosen explicitly is impossible. Therefore, we analyze $k$ nearest instances specified by the actual \ac{knn} algorithm on the feature space after training of ETRAG, which turns out that they are almost the same instances used in ETRAG.

We took statistics on the retrieved instances linked to the extraction target. The targets for statistics are the relation labels, head entities, and tail entities, which are closely related to the relation extraction. 
We calculate the percentage of matches between the retrieved instances and the target of extraction in relation labels and the surface of entities.

\begin{table}[t]
    \centering
    \caption{Comparison of contents between retrieved instances and extraction target. Label means the relation label is the same. Entity means either head entity or tail entity is contained. Related means either head entity, tail entity, or relation label is contained.}
    \begin{tabular}{l|r|rrr|r}
        \hline
         & Label & Head entity & Tail entity & Entity & Related\\
        \hline
        top-1 & 72.2 & 7.1 & 3.4 & 10.0 & 73.6 \\
        top-3 & 67.9 & 8.3 & 4.4 & 12.0 & 77.4 \\
        top-5 & 61.8 & 10.9 & 6.3 & 15.7 & 78.9 \\
        top-10 & 60.4 & 18.6 & 10.9 & 25.8 & 82.1 \\
        \hline
    \end{tabular}
    \label{tab:contain}
\end{table}

The statistics in \cref{tab:contain} show the percentage of the retrieved instances that contain objects related to relation extraction. First, for the Label column, the percentage gradually decreases from top-1 to top-10. This indicates that the feature space of the retrieval is structured based on the type of relation labels and that instances with the same relation labels are placed close to each other. For the Entity column, the percentage gradually increases from top-1 to top-10, indicating that including entities is a second retrieval perspective, although the percentages are less than those in the Label column. The results for the Head and Tail entity columns show that instances containing the head entity are more intensive.
From the results of the top-10 row in the Related column, more than 80\% have been selected that contain labels or entities relevant for relation extraction. This may be due to the use of relation and entity features for retrieval. As a result of end-to-end training from these features, the distance becomes smaller when the relation labels match or entities are included. These properties were not given intentionally but were acquired only by training through end-to-end relation extraction, indicating that the retriever does not require any special training.

\section{Conclusions}
\label{sec:conclusion}

This study introduced a novel approach to the text generation model by implementing a retriever with the differentiable $k$-nearest neighbor selection for end-to-end trainable modeling.
Existing models with retrievers cannot train end-to-end due to the non-differentiable environments of the instance selection part and the integration part of instances. Therefore, we proposed a fully differentiable and end-to-end trainable \ac{rag} ETRAG by differentiable $k$-nearest neighbor selection and integration as a soft prompt.
Our method, centered around neural prompting, significantly enhances the retriever's capability to select instances for use in prompts. 

Experimental findings underscore this approach's effectiveness, particularly in scenarios with limited training data in evaluating relation extraction performance. We evaluated the model with ETRAG and compared the model without ETRAG with existing methods with the TACRED dataset. Our experiments showed that our proposal ETRAG consistently improved from the baseline model without ETRAG. Moreover, the model reported outstanding performance in low-resource settings, especially the new state-of-the-art for the TACRED dataset in the 10\% training data setting.

Our analysis confirmed that the number of retrieved instances introduced by ETRAG can balance the precision-recall trade-off. We also confirmed that the end-to-end trained retriever referred to the instances involved in relation extraction. However, our study also identified limitations in our method's performance when training instances are sufficient. 

Future work could focus on refining the retriever's training process to adapt more effectively to varying sizes of training datasets and exploring ways to optimize instance selection for a broader range of data scenarios. Moreover, since we evaluated on only relation extraction while ETRAG can be applied to other text generation models, further evaluation of other tasks, such as question answering~\cite{kim-etal-2023-tree-rag-qa}, will be conducted to confirm the applicability of ETRAG.

\section*{Acknowledgements}
The authors would like to express gratitude to the members of the Computational Intelligence Laboratory for their helpful comments on this study. This work was supported by Grant-in-Aid for JSPS Fellows Grant Number JP22J14025. Computational resources of AI Bridging Cloud Infrastructure (ABCI) provided by the National Institute of Advanced Industrial Science and Technology (AIST) were used.

\bibliographystyle{IEEEtran}
\bibliography{ref}

\begin{IEEEbiography}[{\includegraphics[width=1in,height=1.25in,clip,keepaspectratio]{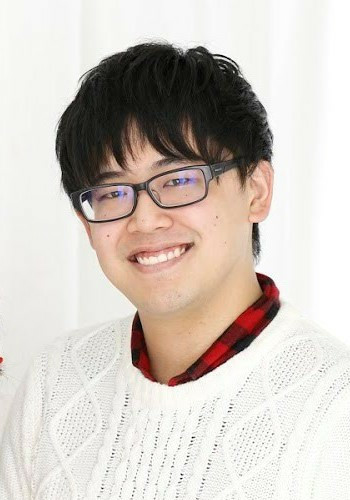}}]{Kohei Makino} received B.E. and M.E. degrees from Toyota Technological Institute, Aichi, Nagoya, Japan, where he is currently pursuing a doctor's degree.
His research interests include deep learning, natural language processing, and information extraction.
\end{IEEEbiography}

\begin{IEEEbiography}[{\includegraphics[width=1in,height=1.25in,clip,keepaspectratio]{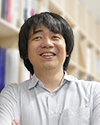}}]{Makoto Miwa} received a Ph.D. degree from the University of Tokyo in 2008. He is currently a professor at the Toyota Technological Institute and an invited researcher at the National Institute of Advanced Industrial Science and Technology. His research interests include natural language processing, deep learning, and information extraction.
\end{IEEEbiography}

\begin{IEEEbiography}[{\includegraphics[width=1in,height=1.25in,clip,keepaspectratio]{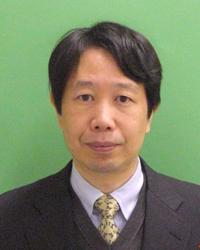}}]{Yutaka Sasaki} received his B.E., M.Eng. and Ph.D. in Engineering from the University of Tsukuba in 1986, 1988, 2000, respectively. From 1988 to 2006, he was with the NTT laboratories, Japan. During 2004-2006, he was a department head at the Advanced Telecommunication Research Institute International (ATR), Kyoto, Japan. 
In 2006, he joined the School of Computer Science, the University of Manchester and the National Center for Text Mining (NaCTeM) in UK. In 2009, he moved to the Toyota Technological Institute, Nagoya, Japan since then he is a Professor at the Department of Advanced Science and Technology. He is also an adjoint professor at the Toyota Technological Institute at Chicago. His recent research interests include machine learning, natural language processing, deep state-space models, and biomedical/materials informatics.
\end{IEEEbiography}

\EOD

\end{document}